\pdfoutput=1
\documentclass[11pt]{article}
\usepackage[final]{acl}
\usepackage{times}
\usepackage{latexsym}
\usepackage[T1]{fontenc}
\usepackage[utf8]{inputenc}
\usepackage{graphicx}
\usepackage{booktabs}
\usepackage{amsmath}
\usepackage{amssymb}
\usepackage{array}
\usepackage{multirow}
\usepackage{xcolor}
\usepackage{url}
\usepackage{pifont}
\usepackage{dblfloatfix}
\newcommand{\cmark}{\textcolor{green!60!black}{\ding{51}}}
\newcommand{\xmark}{\textcolor{red}{\ding{55}}}
\newcommand{\pmark}{\textcolor{orange!80!black}{\ding{124}}}

\title{NSIDDx: A Design Framework for Neuro-Symbolic, Practitioner-First\\
Differential Diagnosis in Low-Resource Settings}

\author{
  \textbf{Aarav Singh} \\
  IIIT Naya Raipur \\
  \texttt{aarav24101@iiitnr.edu.in}
}

\begin{document}
\maketitle

\begin{abstract}
LLM-based diagnostic systems achieve high semantic accuracy on benchmarks,
but open-ended evaluation on clinically uncommon presentations reveals a
systematic gap between headline accuracy and verifiable clinical reliability.
We evaluate an LLM+rare-disease-RAG pipeline across two cohorts and show
that the paradigm produces confident outputs that are frequently unverifiable
and systematically resistant to clinician interrogation.
We present \textbf{NSIDDx} (Neuro-Symbolic Integrated Differential Diagnosis System), a design framework arguing that DDx systems in
low-resource settings must treat the clinician as an active reasoning agent.
We instantiate this through a neuro-symbolic pipeline with ternary symptom
encoding, contradiction detection, audit strings, and practitioner override
--- running offline on consumer hardware. We distill five design principles
for clinician-in-the-loop clinical NLP and invite the prospective studies
needed to validate the claim at scale.
\end{abstract}

\section{Introduction}
\label{sec:introduction}

Differential diagnosis is among the most cognitively demanding tasks in
clinical medicine. A general practitioner must retrieve relevant disease
knowledge, weigh present and absent symptoms, consider rare conditions,
and produce a reasoned conclusion --- often in under fifteen minutes without
specialist support. In low-resource settings, this load falls on a single
practitioner frequently without any decision support tool. They do not need
a system that is right eighty percent of the time if they cannot tell which
twenty percent to distrust. They need a \textit{practitioner-first} design
whose reasoning they can follow, augment, and whose uncertainty is surfaced
rather than concealed.

Large language models have shown strong performance on clinical benchmarks
\citep{singhal2023large,nori2023gpt4}, yet the clinician receiving an
LLM-based diagnostic output gets an answer, not a reasoning process
they can interrogate, modify, or reject on clinical grounds --- a
limitation that persists even in KG-grounded pipelines
\citep{chandak2023primekg,gargano2024hpo,wang2025kgllm} where the graph informs
generation but remains invisible to the clinician. \citet{singhal2023large} themselves note that strong benchmark
numbers conceal key gaps when models are evaluated by clinicians rather than
automated metrics --- a finding our open-ended evaluation on 750 CUPCase cases
independently reproduces: 75.2\% semantic accuracy at DDx@5 collapses to
6.6\% exact match, and 83.5\% of rare scanner outputs carry no tokenically
verifiable (verifiable by words (token) overlapping between disease names) relationship to the ground-truth diagnosis. The clinical decision
support literature consistently identifies automation bias and alert fatigue
as major barriers to adoption \citep{goddard2012automation}.
This is not a failure of accuracy; it is a failure of design philosophy.

\paragraph{Our Position.}
We argue that for low-resource clinical settings, the most critical design
constraint for diagnostic AI is not accuracy alone, but
\textit{tractability of disagreement}: a system whose reasoning is
transparent, auditable, and modifiable by the clinician may in practice be
more useful than a more accurate system whose failures are invisible ---
because it invites collaboration rather than anchoring.

This paper presents \textbf{NSIDDx}, a neuro-symbolic differential diagnosis
pipeline built around that philosophy. NSIDDx uses a large language model for
what it does well --- synthesising clinical narratives, generating structured
reasoning chains, producing readable explanations --- while independently
validating that reasoning through a symbolic layer grounded in HPO/MONDO
\citep{gargano2024hpo} and PrimeKG \citep{chandak2023primekg}. The practitioner can read the Propositional Logic (PL) audit
string, inspect the graph, and disagree: they can add a symptom the model
missed, remove a candidate they consider implausible, and rerun the scoring
pipeline. The loop is not closed without them.

Our contributions are two-fold:

\textbf{(A) Position and Design Principles.} We distill five design
principles for clinician-in-the-loop clinical NLP --- surface contradiction,
preserve negation structurally, make override cheap, provide multi-level
explanation, and design for offline deployment --- intended as actionable
guidelines for the community.

\textbf{(B) System Instantiation and Failure Characterisation.} We present
NSIDDx as a concrete instantiation of these principles, built around a
ternary symptom encoding scheme ($+1$ present, $-1$ explicitly denied, $0$
unrecorded), a dual-formula scoring engine combining a
contradiction-sensitive matrix score and a coverage-penalising hybrid score,
and a Negation-Retaining Phenotype Graph that renders absent findings as
structural nodes rather than silently discarding them. We evaluate this pipeline under open-ended conditions with no multiple-choice
scaffolding, characterising systematic failure modes --- vocabulary
mismatch, confidence-without-grounding, and negation inversion --- that make
clinician intervention a routine requirement rather than an edge case.
NSIDDx's symbolic transparency layer is proposed as the architectural
response. We do not claim diagnostic superiority --- our contribution is
architectural.

We acknowledge that no user study or clinical validation has been conducted.
The argument for clinician involvement is theoretical and demonstrated by
one case study, and explicitly invites prospective validation.

\section{Related Work}
\label{sec:related}

\paragraph{LLM-based clinical diagnosis.}
Large language models have demonstrated strong performance on medical
benchmarks \citep{singhal2023large,nori2023gpt4}. Health-LLM
\citep{yu2025healthllmpersonalizedretrievalaugmenteddisease} extends this with a two-pass RAG pipeline and
XGBoost classifier for personalised prediction, but the clinician receives
a label with no inspectable logic chain and no mechanism for override.

\paragraph{Human-AI collaborative diagnosis.}
AMIE \citep{tu2025amie} shows conversational AI can match physician accuracy
in OSCE studies but does not expose an inspectable reasoning chain or
practitioner override mechanism. In low-resource settings, the ability to
interrogate and correct AI suggestions may be more valuable than raw accuracy.

\paragraph{Knowledge graph-guided iterative diagnosis.}
MedKGI \citep{wang2025medkgi} models multi-step diagnosis as an iterative
conversational loop using information gain over KG subgraphs. It does not
model absent symptoms --- negative findings are neither elicited nor encoded,
and clinicians cannot see which absent symptoms are relevant to the
differential. NSIDDx is complementary: where MedKGI optimises automated
history-taking over positive findings, NSIDDx makes negation explicit and
inspectable after history collection.

\paragraph{Knowledge graph grounding for prognosis.}
\citet{wang2025kgllm} fine-tune lightweight LLMs on PrimeKG-scaffolded
reasoning paths for next-visit prediction, encoding visits as binary ICD-9
vectors --- structurally unable to represent explicit symptom denials
($x_i = -1$). The reasoning chain is not inspectable and absent findings are
implicit. Our two-cohort evaluation quantifies this opacity: 41.6\% of DDx@5
semantic hits on CUPCase are tokenically unverifiable without an audit trail.

\paragraph{Neuro-symbolic explainability.}
\citet{mondal2025ltn} and \citet{lu2024neurosymbolic} apply LNN-based
methods to explainable diabetes prediction, exposing auditable reasoning with
negation support, but operate on low-dimensional tabular data with statically
defined rules and cannot parse unstructured clinical narratives.

\paragraph{Rare disease detection.}
RADAR \citep{radar2025} applies FAISS to rare disease retrieval in brain
MRI; \citet{rare2024kg} explore knowledge-guided RAG for rare disease
diagnosis. NSIDDx's scanner differs: it operates over clinical text, is
grounded in ZebraMap \citep{islam2025zebramap}, and scores candidates
through the same dual-source symbolic pipeline as the primary DDx, offline
without API access.

\paragraph{White-box diagnostic systems.}
Ada DX \citep{ronicke2019adadx} demonstrates the value of accepting present
and absent findings with full practitioner override, but operates as a
probabilistic expert system on structured symptom input rather than
free-text narratives.

\paragraph{Feature comparison.}
Table~\ref{tab:comparison} situates NSIDDx among prior systems. The claim
is not individual feature novelty but that the \textit{specific integration}
of all five dimensions --- negation modelling, override interface, offline
capability, unstructured text input, and KG grounding --- is novel in
neuro-symbolic clinical NLP.

\begin{table}[t]
\footnotesize
\centering
\setlength{\tabcolsep}{2.5pt}
\renewcommand{\arraystretch}{0.92}
\begin{tabular}{@{}lp{0.28cm}p{0.28cm}p{0.28cm}p{0.28cm}p{0.28cm}@{}}
\toprule
\textbf{System} & \textbf{N} & \textbf{O} & \textbf{C} & \textbf{U} & \textbf{K} \\
\midrule
Health-LLM \citep{yu2025healthllmpersonalizedretrievalaugmenteddisease}              & \xmark & \xmark & \xmark & \cmark & \cmark \\
MedKGI \citep{wang2025medkgi}                   & \xmark & \xmark & \xmark & \cmark & \cmark \\
LNN \citep{mondal2025ltn,lu2024neurosymbolic}   & \cmark & \xmark & \cmark & \xmark & \xmark \\
Ada DX \citep{ronicke2019adadx}                 & \cmark & \cmark & \cmark & \xmark & \cmark \\
AMIE \citep{tu2025amie}                         & \xmark & \xmark & \xmark & \cmark & \xmark \\
\midrule
\textbf{NSIDDx}                                 & \pmark & \cmark & \cmark & \cmark & \cmark \\
\bottomrule
\end{tabular}
\caption{Feature comparison. Columns: \textbf{N}egation modelling,
  \textbf{O}verride interface, offline \textbf{C}apable,
  \textbf{U}nstructured text input, \textbf{K}G grounding.
  \cmark~full; \pmark~partial (feature exists but with major constraints:
e.g., negation requires manual entry, override is single-direction, or
offline mode excludes some functionality); \xmark~none.}
\label{tab:comparison}
\end{table}

\section{System Architecture}
\label{sec:architecture}

\begin{figure*}[t]
\centering
\includegraphics[width=0.85\textwidth]{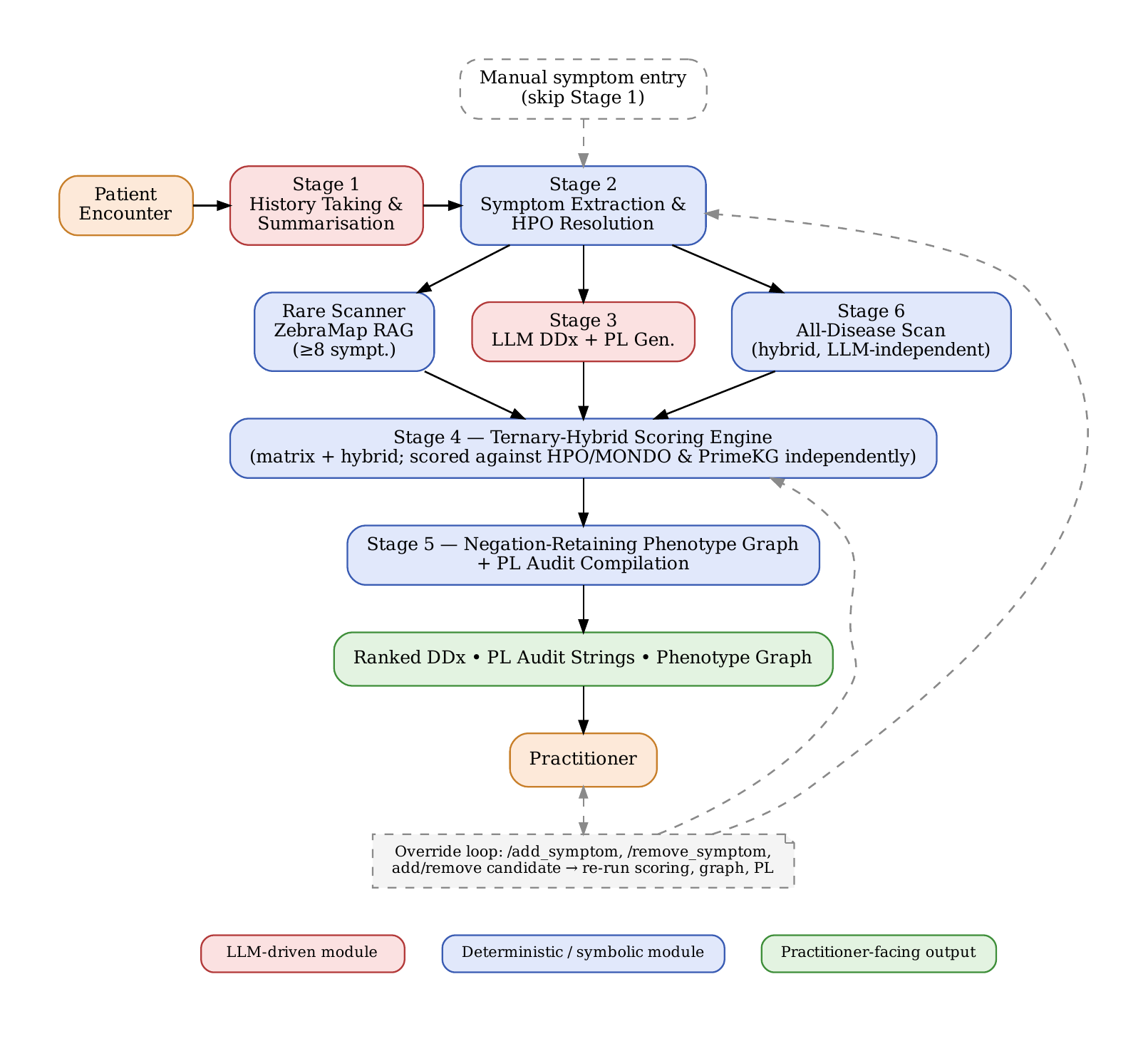}
\caption{NSIDDx system architecture (colour key in figure). Solid arrows trace the path exercised in the evaluation of Section~\ref{sec:eval}: history taking (Stage~1) feeds symptom extraction (Stage~2), whose ternary-encoded output is passed in parallel to the rare disease scanner, the LLM differential diagnosis module (Stage~3), and the all-disease scanner (Stage~6); all three candidate streams are scored by the ternary-hybrid engine (Stage~4) before graph and audit-string generation (Stage~5). Dashed arrows show that the pipeline is not rigid: the practitioner may enter symptoms manually in place of Stage~1, and the override loop lets the practitioner add or remove symptoms (fed back into Stage~2) or candidates (fed back into Stage~4) at any point, triggering re-scoring, re-graphing, and PL recompilation (Section~\ref{sec:architecture}, ``Practitioner override'').}
\label{fig:architecture}
\end{figure*}

NSIDDx is a modular pipeline with six core stages (Figure~\ref{fig:architecture}). The pipeline is not rigidly sequential: every stage is an independently invocable module, and the practitioner can bypass Stage~1 entirely by entering symptoms directly, or add and remove symptoms and candidates at any later point via the override interface, which re-triggers scoring, graph generation, and PL compilation without restarting the pipeline. Figure~\ref{fig:architecture} distinguishes the path exercised in the evaluation of Section~\ref{sec:eval} (solid arrows) from these optional bypass and override affordances (dashed arrows).

\paragraph{Data sourcing.}
Knowledge base files are sourced from HPO (hp.obo, phenotype.hpoa), MONDO
(mondo.owl), and PrimeKG (kg.feather). HPO terms are extracted into a
symptom catalog; unannotated modifier and inheritance terms are excluded.
MONDO cross-references unify OMIM and ORPHA identifiers; entries with zero
present-symptom annotations are dropped, yielding roughly 10,800 disease
profiles. For PrimeKG, disease-phenotype edges are bridged to HPO IDs;
disease-disease edges are excluded. Over 9,000 phenotype-bearing entities
enter the scoring pipeline. The NSIDDx pipeline, including all preprocessing
scripts and filter thresholds, is publicly available at
\url{https://github.com/joetheguide2/NSIDDX-}.

\subsection{Stage 1: Clinical History Taking and Summarisation}

Clinical history is gathered via an LLM-driven conversational agent and
condensed into a structured summary preserving both positive and negative findings.

\subsection{Stage 2: Symptom Extraction and HPO Resolution}

Symptom extraction passes the clinical summary through a two-step pipeline.
First, the LLM reformats the summary into a structured format with
\texttt{Presence of:} and \texttt{Denies:} prefixes. Second, surface forms are resolved to HPO IDs via MedSpaCy TargetRule
(literal) or Abhinand/MedEmbed-large-v0.1 (semantic) matching; each
is tagged with its resolution method. The result is two typed sets: \texttt{patient\_hpo\_ids}
(present) and \texttt{absent\_hpo\_ids} (explicitly denied).

Disease name resolution uses exact then substring match, preferring
diseases with symptom profiles over umbrella terms like ``syndrome''.

\subsection{Stage 3: LLM Differential Diagnosis and PL Generation}

The LLM differential diagnosis stage sends the clinical summary together
with the confirmed present and explicitly denied symptom lists to the LLM
under a structured DDx prompt. The prompt enforces a fixed
\texttt{Evidences $\rightarrow$ Reasoning $\rightarrow$ Diagnosis} format
per entry, with an \texttt{ABSENT:} marker for denied symptoms.

Propositional Logic (PL) strings — case-specific conjunctions of symptom
propositions implying a diagnosis — are generated per diagnosis, either
symbolically (from HPO-matched symptoms) or via the LLM. Propositional
statements use absent symptoms as negated propositions (e.g., $\neg(\text{Raynaud
phenomenon})$). These strings serve as practitioner-readable audit trails
for each diagnostic candidate.

\subsection{Stage 4: Ternary-Hybrid Scoring Engine}

The scoring engine runs two scoring formulas — the \textit{matrix score}
and the \textit{hybrid score} — independently against each knowledge source
(HPO/MONDO, PrimeKG), yielding independent scores per candidate disease.
Both formulas are applied to all sources.

\paragraph{Score Formula 1: Matrix Score.}
A patient symptom vector $\mathbf{V}_p \in \{-1, 0, +1\}^N$ is built from
the extracted HPO IDs: $+1$ for confirmed present, $-1$ for explicitly
denied, $0$ for unrecorded. A disease profile vector
$\mathbf{A}_j \in \{-1, 0, +1\}^N$ encodes expected phenotypes. The score
is the normalised dot product:
\[
S_j^{\text{matrix}} = \frac{\mathbf{V}_p \cdot \mathbf{A}_j}{|\mathbf{A}_j|}
\in [-1,\, +1]
\]
where $|\mathbf{A}_j|$ is the disease's profile size. Negative scores
indicate contradictions --- the patient denies a required symptom or
presents one the disease expects absent. Both formulas use standard measures (dot product,
Jaccard); their specific combination with equal weights is novel.

\paragraph{Score Formula 2: Hybrid Score.}
The hybrid score measures how well a disease explains the patient's
positive findings:
\[
S_j^{\text{hybrid}} = 0.5 \times \frac{|A \cap B|}{|A|}
    + 0.5 \times \frac{|A \cap B|}{|A \cup B|}
\in [0,\, +1]
\]
Here $A$ is the patient's set of confirmed present symptoms and $B$ is
the disease's expected phenotype profile (the positive-symptom
counterparts of $\mathbf{V}_p$ and $\mathbf{A}_j$ above, with denied
and unrecorded symptoms excluded). The inclusion term ($|A \cap B|/|A|$) rewards covering most of the
patient's symptoms; the Jaccard term ($|A \cap B|/|A \cup B|$) penalises
large profiles with only incidental symptom overlap.

\subsection{Stage 5: Negation-Retaining Phenotype Graph and Rare Disease RAG Scanner}

The negation graph stage builds a \textbf{Negation-Retaining Phenotype
Graph} — a directed multigraph using NetworkX from the current patient
payload and the last scoring output. The graph contains five node types:
present symptoms (HPO-resolved), absent symptoms (explicitly denied),
past medical history, family history, and DDx candidates.
Absent symptom nodes remain structurally present in the graph — visually
distinguishable from confirmed symptoms — giving the practitioner a direct
view of pertinent negatives alongside positive findings.

\paragraph{Edge types.}
The graph renders four edge types from PrimeKG: \texttt{explains}
(disease $\rightarrow$ symptom); \texttt{presents} (symptom $\rightarrow$
disease); \texttt{linked\_to} (symptom co-occurrence); and
\texttt{phenotype\_modifier\_of} (modifier $\rightarrow$ base).
\texttt{Explains} edges carry a PrimeKG-derived specificity weight
$w = \log_{10}(\text{total diseases}/\text{symptom count}) + 1$,
where higher weight indicates greater diagnostic specificity. The scoring
layer treats modifier terms as independent phenotypes — a known limitation
(Section~\ref{sec:limitations}).

The rare disease RAG scanner provides an optional safety net grounded in
ZebraMap \citep{islam2025zebramap}, encoded with MedEmbed and indexed in a
FAISS vector store. Retrieved candidates are mapped from UMLS to MONDO IDs,
enabling scoring through the same dual-source symbolic pipeline as the
primary DDx. The module is intended for presentations with eight or more
symptoms where the primary DDx may miss low-prevalence conditions.

\subsection{Stage 6: All-Disease Scanning}

Beyond scoring LLM-nominated candidates, NSIDDx provides a parallel,
LLM-independent diagnostic pathway that sweeps the entire knowledge base,
scoring every disease in HPO/MONDO and PrimeKG against the current
patient symptom set using the hybrid score.
The hybrid formula is used by deliberate design: in full-database sweeps,
the matrix score suffers a structural bias where diseases with tiny profiles
achieve artificially high scores (Table~\ref{tab:bias}).

\begin{table}[ht]
\small
\centering
\begin{tabular}{@{}lccc@{}}
\toprule
\textbf{Disease} & \textbf{Matrix} & \textbf{Hybrid} & \textbf{Explained} \\
\midrule
Intellectual dev.\ disorder & \textbf{1.000} & 0.071 & 1/14 \\
17q11.2 microduplication    & 0.625          & \textbf{0.762} & 12/14 \\
\bottomrule
\end{tabular}
\caption{Matrix score bias in full-database scanning. The hybrid score
  correctly reverses the ranking, which is why all full-database sweeps
  use the hybrid formula by default.}
\label{tab:bias}
\end{table}

\paragraph{Practitioner override and closed-loop design.}
At any stage, the practitioner can add or remove symptoms and disease
candidates, then re-run scoring, graph generation, and PL compilation.
Changes propagate back into the LLM prompt, giving direct control over
the differential diagnosis without modifying underlying weights.

\section{Evaluation}
\label{sec:eval}

This section evaluates the LLM+rare-disease-RAG diagnostic pipeline on
750 CUPCase cases across two cohorts, characterising failure modes of the
paradigm rather than of NSIDDx alone.

\subsection{Setup}

We evaluate on two cohorts drawn from \textbf{CUPCase}
\citep{cupcase2025}, a publicly available benchmark of 3,562 real-world
patient case reports sourced from BMC case report journals, accepted at
AAAI 2025. The \textbf{500-case random sample} (CUP) represents the full
distribution of clinically uncommon presentations --- the realistic
evaluation surface for any system deployed on edge-case presentations.
The \textbf{250-case exact-match sample} (WELL) consists of cases whose
correct diagnosis resolves by exact name match in HPO/MONDO or PrimeKG;
this cohort represents the upper bound of automated symbolic pipeline
performance within CUPCase, as the correct disease is guaranteed to have
an ontology entry, a curated phenotype profile, and a resolvable name.
Approximately 700 of 3,562 CUPCase cases meet this criterion. Both
cohorts are drawn from CUPCase; there is no domain shift between them.
The only variable is ontology coverage. All evaluation used Qwen3.5-9B \citep{qwen3.5} (IQ4\_XS, a 4-bit GGUF quantization) on consumer hardware without cloud API access. Median per-case runtime was 87.6~seconds.
\paragraph{Metrics.}
We evaluate under four matching metrics of increasing permissiveness:
\textbf{exact} (complete string match), \textbf{substring} (one is a
complete substring of the other), \textbf{token} (significant token
overlap), and \textbf{semantic} (cosine similarity of sentence embeddings using Abhinand/MedEmbed-large-v0.1, similarity threshold 0.7).
We report all four to make the gap between verifiable and apparent
accuracy explicit.

\paragraph{Statistical analysis.}
Accuracy differences between cohorts are tested using the chi-square test
of proportions on 2$\times$2 contingency tables (hit/miss per case);
all reported expected cell counts exceed five. Continuous distributions
(phenotype counts, DDx label lengths, confidence scores) are compared
using the two-sided Mann-Whitney U test, which makes no normality
assumption. Per-cohort confidence intervals are Wilson score intervals
at $\alpha = 0.05$. Significance markers follow the convention
$^{*}p<0.05$, $^{**}p<0.01$, $^{***}p<0.001$.

\subsection{Primary DDx Accuracy}

\begin{table}[ht]
\small
\centering
\setlength{\tabcolsep}{4pt}
\begin{tabular}{lcccc}
\toprule
\textbf{Pipeline} & \textbf{Exact} & \textbf{Token} & \textbf{Semantic} \\
\midrule
\multicolumn{4}{c}{\textbf{CUP (500-case random sample)}} \\
DDx (LLM) @1  & 2.4\%  & 17.0\% & 41.8\% \\
DDx (LLM) @3  & 5.6\%  & 29.2\% & 67.8\% \\
DDx (LLM) @5  & 6.6\%  & 34.4\% & \textbf{75.2\%} \\
MONDO @5      & 0.0\%  & 1.8\%  & 5.0\% \\
KG @5         & 0.0\%  & 1.4\%  & 4.0\% \\
Rare scanner any & 2.6\% & 9.4\% & 23.6\% \\
DDx@5 $\cup$ Rare & --- & ---    & 78.4\% \\
\midrule
\multicolumn{4}{c}{\textbf{WELL (250-case exact-match sample)}} \\
DDx (LLM) @1  & 7.2\%  & 25.6\% & 42.8\% \\
DDx (LLM) @3  & 11.6\% & 40.8\% & 66.4\% \\
DDx (LLM) @5  & 14.0\% & 46.8\% & \textbf{73.6\%} \\
MONDO @5      & 0.4\%  & 2.8\%  & 8.0\% \\
KG @5         & 0.8\%  & 5.6\%  & 8.0\% \\
Rare scanner any & 17.6\% & 27.6\% & 35.6\% \\
DDx@5 $\cup$ Rare & ---  & ---    & 80.4\% \\
\bottomrule
\end{tabular}
\caption{DDx semantic accuracy is statistically equivalent between
  cohorts ($\chi^2$, $p=0.700$). Exact and token accuracy diverge
  significantly ($p<0.001$--$0.01$), reflecting vocabulary mismatch on
  uncommon cases rather than differential LLM reasoning. CUP = 500-case
  random sample; WELL = 250-case exact-match sample (upper bound of
  ontology coverage within CUPCase).}
\label{tab:accuracy}
\end{table}

The LLM component performs equivalently across both cohorts: DDx@5
semantic accuracy is 75.2\% and 73.6\% respectively ($\chi^2$,
$p=0.700$). The divergence appears in exact accuracy (6.6\% vs 14.0\%,
$p=0.001$) and token accuracy (34.4\% vs 46.8\%, $p<0.001$) --- metrics
that depend on vocabulary alignment between the LLM's generated labels
and ontology entries. Phenotype extraction rates are also statistically
indistinguishable (MWU, $p=0.976$), ruling out differential input quality
as a confound. The failure is at the vocabulary interface, not in the
LLM's clinical reasoning. The MONDO and KG symbolic scorers achieve
5.0--8.0\% semantic accuracy in both cohorts, confirming that their
near-zero performance reflects HPO phenotype annotation sparsity, not a
scoring formula failure.

\subsection{Rare Disease Scanner}

The rare scanner recovers cases missed by the LLM DDx (Table~\ref{tab:rare}).

\begin{table}[ht]
\small
\centering
\setlength{\tabcolsep}{3pt}
\begin{tabular}{p{3.1cm}c}
\toprule
\textbf{Metric} & \textbf{CUP / WELL} \\
\midrule
Empty-match rate & 83.5\% / 84.4\% \\
Mean empty-match confidence & 0.641 / 0.645 \\
Discriminative gap (correct vs missed) & +0.061*** / +0.064*** \\
\midrule
Rare scanner sole recovery & 3.2\% / 6.8\% \\
DDx@5 $\cup$ Rare (combined ceiling) & 78.4\% / 80.4\% \\
\bottomrule
\end{tabular}
\caption{Rare disease scanner quality analysis. The empty-match rate is
  consistent across cohorts. The discriminative gap is statistically
  significant (MWU, $p<0.001$) but clinically insufficient as a
  decision threshold.}
\label{tab:rare}
\end{table}

Candidates are ranked by embedding similarity without phenotypic pathway
validation; the scanner functions as a hypothesis generator requiring
clinician review.

\subsection{Accuracy by Extraction Quality}

\begin{table}[ht]
\small
\centering
\setlength{\tabcolsep}{4pt}
\begin{tabular}{lcccc}
\toprule
\textbf{Phenotype bin} & \multicolumn{2}{c}{\textbf{CUP}} & \multicolumn{2}{c}{\textbf{WELL}} \\
\cmidrule(lr){2-3} \cmidrule(lr){4-5}
 & $n$ & \textbf{DDx@5} & $n$ & \textbf{DDx@5} \\
\midrule
0 (extraction failure) & 12 & 83.3\%* & 5 & 60.0\%* \\
1--2 & 96 & 74.0\% & 54 & 66.7\% \\
3--5 (modal) & 173 & 72.3\% & 84 & 75.0\% \\
6--9 & 131 & 79.4\% & 60 & 75.0\% \\
10+ & 88 & 75.0\% & 47 & 78.7\% \\
\bottomrule
\end{tabular}
\caption{DDx semantic accuracy by extraction quality. No between-cohort
  difference is significant at any phenotype bin (all $p>0.44$). The
  absence of significant differences within bins confirms that performance
  divergence is attributable to the vocabulary boundary layer, not
  differential extraction quality.}
\label{tab:extraction}
\end{table}

Extraction errors are input-level failures that the clinician can
correct regardless of ontology coverage — motivating the practitioner
override interface.

\section{Qualitative Failure Mode Analysis}
\label{sec:failures}

Analysis of automated failure modes across both cohorts reveals four
intervention categories (Table~\ref{tab:taxonomy}).

\begin{table}[ht]
\small
\centering
\setlength{\tabcolsep}{4pt}
\begin{tabular}{@{}p{3.2cm}cc@{}}
\toprule
\textbf{Category} & \textbf{CUP (500)} & \textbf{WELL (250)} \\
\midrule
Cat 1: Extraction failure & 2.4\%  & 2.0\% \\
Cat 2: DDx complete miss  & 24.4\% & 25.6\% \\
Cat 3: Semantic-only hit  & 41.6\% & 28.0\%*** \\
Cat 4: KG+MONDO silent    & 70.0\% & 62.8\% \\
\midrule
\textbf{Any intervention} & \textbf{96.8\%} & \textbf{93.2\%}* \\
\bottomrule
\end{tabular}
\caption{Failure taxonomy across both cohorts. Categories 1 and 2 are
  statistically equivalent ($p>0.78$), indicating LLM blind spots and
  extraction failures are domain-level phenomena not dependent on
  ontology coverage. Category 3 diverges significantly ($p<0.001$),
  localising the primary performance gap to the vocabulary boundary.}
\label{tab:taxonomy}
\end{table}

\paragraph{Negation Inversion.}
Analysis of HPO resolver mappings identified 26 cases where explicitly
denied symptoms were semantically mapped to affirmative HPO terms,
entering the scoring pipeline as positive evidence. Examples include:
\texttt{DENIED: pain} $\rightarrow$ Pain insensitivity, \texttt{DENIED:
hemoptysis} $\rightarrow$ Hemoptysis, \texttt{No remarkable family history}
$\rightarrow$ hereditary fructose intolerance. In the sarcoidosis case
(Section~\ref{sec:case}), six of seven absent symptoms were LLM-inferred
rather than explicitly denied, causing matrix score $-0.024$ for the
correct diagnosis. This failure class is structurally undetectable in any
pipeline without explicit polarity encoding.

\paragraph{Implications for design.}
Categories 1 and 2 are domain-level failures requiring clinician
oversight regardless of ontology coverage. Category 3 calls for synonym
normalisation and an audit trail so the clinician can verify tokenically
unverifiable hits. Category 4 confirms that symbolic path confirmation
is unavailable for most cases; the PL string and negation graph provide
alternative explanation modalities independent of score magnitude.
Active contradictions and negation inversions require no system change
--- surfacing them is the design goal.

\section{Case Study}
\label{sec:case}

The following case falls within the category of semantic-only hit under
automation with active symbolic contradiction and illustrates the
override mechanism converting a surfaced failure into a confirmed
diagnosis.

\paragraph{Case and automated extraction failure.}
A 64-year-old Japanese woman presents with exertional dyspnea, bilateral
pleural effusions, bilateral hilar and mediastinal lymphadenopathy,
subcutaneous nodules, and non-caseous epithelioid granulomas on biopsy.
The correct diagnosis is Sarcoidosis
\citep{kesici2014sarcoidosis}.
The automated pipeline extracts 5 present and 7 absent symptoms:

Present (5): \texttt{HP:0002094} dyspnea, \texttt{HP:0032252} granuloma,
\texttt{HP:0034388} hilar lymphadenopathy, \texttt{HP:0100721} mediastinal
lymphadenopathy, \texttt{HP:0002202} pleural effusion.

Absent (7): \texttt{HP:0100749} $\neg$chest pain, \texttt{HP:0012735}
$\neg$cough, \texttt{HP:0001945} $\neg$fever, \texttt{HP:0002105}
$\neg$hemoptysis, \texttt{HP:0030166} $\neg$night sweats, \texttt{HP:0001962}
$\neg$palpitations, \texttt{HP:0001824} $\neg$weight loss.

All seven absent symptoms are generated by the LLM's structured reformatting
step, which infers pertinent negatives from the narrative even where the case
text does not explicitly deny them. Splenomegaly, documented via gallium-67
scintigraphy showing abnormal splenic uptake, is not extracted because no
explicit ``splenomegaly'' surface form appears in the HPO synonym dictionary.

The resulting automated score for Sarcoidosis is
$\text{MONDO}_m = -0.024$, $\text{MONDO}_h = 0.512$ --- the negative matrix
score places it last in the symbolic ranking despite the LLM correctly
nominating it first in the DDx. The system does not suppress this
contradiction; the negative score is surfaced in the scoring output
alongside the PL audit trail.

\paragraph{Human-in-the-loop demonstration.}
A researcher reviewed the raw case narrative alongside the automated symptom
vector and performed three targeted interventions:

\begin{enumerate}
\item Corrects the absent list: six of the seven absent symptoms (chest
  pain, cough, fever, haemoptysis, night sweats, weight loss) were not
  explicitly denied in the case text and are removed, retaining only
  $\neg$palpitations (\texttt{HP:0001962}) as a true absent finding.
\item Adds splenomegaly by medical judgement: the gallium-67 scintigraphy
  finding of abnormal splenic uptake implies splenic involvement, even
  though no explicit ``splenomegaly'' string appears in the symptom
  extraction pass. The clinician enters it as
  \texttt{/add\_symptom splenomegaly}.
\item Re-runs \texttt{/score}.
\end{enumerate}

Removing the six spurious absent symptoms resolves the matrix contradiction
immediately: Sarcoidosis moves from $\text{MONDO}_m = -0.024$ to
$\text{MONDO}_m = 0.073$, $\text{MONDO}_h = 0.284$. Adding splenomegaly
improves coverage further: $\text{MONDO}_m = 0.098$,
$\text{MONDO}_h = 0.331$. Sarcoidosis rises to rank 1 in the scored DDx.

The corrected present vector contains 7 HPO IDs (dyspnea, hilar
lymphadenopathy, mediastinal lymphadenopathy, pleural effusion,
subcutaneous nodules, granuloma, splenomegaly) and 1 absent
($\neg$palpitations). The full-database sweeps independently confirm
Sarcoidosis at top ranks across all scoring sources.

The PL audit string for the corrected state reads:

\begin{quote}
(Dyspnea) $\wedge$ (Mediastinal lymphadenopathy) $\wedge$ (Pleural effusion)
$\wedge$ (Splenomegaly) $\Rightarrow$ sarcoidosis, susceptibility to, 1
\end{quote}

\paragraph{Phenotypic graph.}

\begin{figure*}[ht]
    \centering
    \includegraphics[width=\textwidth]{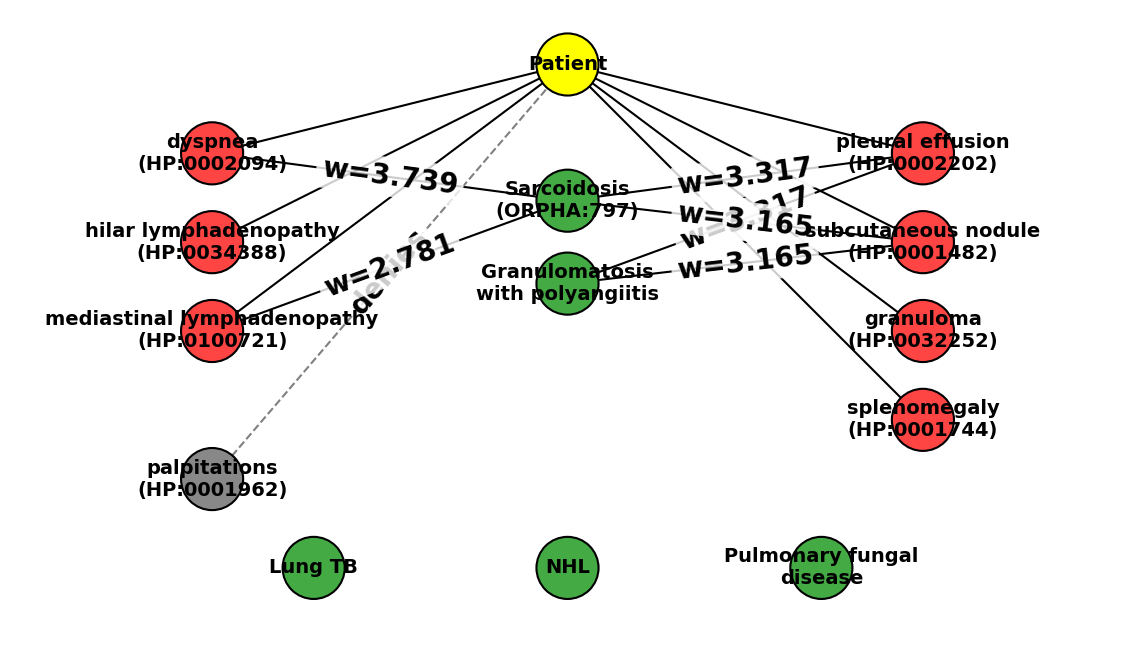}
    \label{fig:graph}
\caption{Negation-Retaining Phenotype Graph for the corrected case.
Red nodes: confirmed present symptoms. Grey node: explicitly denied symptom.
Green nodes: DDx candidates.}
\label{fig:negation_graph}
\end{figure*}

Figure~\ref{fig:negation_graph} shows the Negation-Retaining Phenotype
Graph for the corrected state. 

\textbf{This case demonstrates} the design philosophy: the system surfaced a symbolic
contradiction, the practitioner corrected extraction errors via override,
and convergent evidence confirmed the corrected diagnosis. It does not
represent typical recovery rates (Section~\ref{sec:failures}). The three
interventions required represent capabilities that Category 3 analysis
indicates 208 CUPCase cases would benefit from. We do not claim this
recovery rate is generalisable without a user study; we demonstrate that
the mechanism functions as designed in this instance.
\section{Position: Five Design Principles for Clinician-in-the-Loop Clinical NLP}
\label{sec:position}
From the NSIDDx experience, we distill five design principles for clinical
NLP systems that treat the practitioner as an active reasoning agent rather
than a passive consumer of model outputs. These principles are derived from
the failure modes we observed (Section~\ref{sec:failures}) and the recovery
patterns demonstrated in the case study (Section~\ref{sec:case}). They are
well-motivated hypotheses, not empirically validated claims; prospective
clinician studies are needed to test their usability and generalisability.

\paragraph{Principle 1: Surface Contradiction.}
When the symbolic layer disagrees with the LLM, the system should
present both outputs side-by-side. The contradiction \textit{is} the
signal: it flags uncertainty, extraction failure, or missing clinician
knowledge. In the sarcoidosis case (Section~\ref{sec:case}),
the negative matrix score ($-0.024$) correctly identified an extraction
error --- the system flagged its own mistake rather than hiding it.

\paragraph{Principle 2: Preserve Negation Structurally.}
Denied symptoms are first-class citizens in scoring, graphs, and
reasoning chains. NSIDDx encodes them as $-1$ in the ternary vector
and renders them as grey nodes in the negation graph, visually distinct
from confirmed findings. The sarcoidosis contradiction (matrix $-0.024$
from spurious absent symptoms) motivates this design.

Yet a symptom may be \textit{structurally absent} (the profile expects
it) without being \textit{clinically meaningfully absent} at a given
stage of presentation. The ternary vector cannot represent this
temporal distinction; the override interface exists to admit the
clinical judgment that no encoding can capture by design.
\paragraph{Principle 3: Enable Auditable Practitioner Override.}
The practitioner can add, remove, or modify symptoms at any stage, with
changes propagating through scoring, graph, and reasoning chains in real
time. Override actions are auditable: what changed and why is visible.
In NSIDDx, \texttt{/add\_symptom} and \texttt{/remove\_symptom}
trigger immediate re-scoring and regeneration of all outputs.
\paragraph{Principle 4: Provide Multiple Levels of Explanation.}
Different clinicians prefer different explanation formats: some read PL
audit strings, others inspect graphs, others examine raw scores. NSIDDx
provides all three. No single explanation format works for all users, and
the system should not force a choice. This aligns with the layered
explanation paradigm established in the XAI literature
\citep{doshi2017towards}. Numerical scores alone can be insufficient;
the PL string and negation graph provide complementary modalities
independent of score magnitude.
\paragraph{Principle 5: Design for Offline Deployment on Consumer Hardware.}
Low-resource settings cannot rely on cloud APIs. Systems designed for
these settings must run locally on commodity hardware. This is not a
technical constraint --- it is a design requirement that shapes every
architectural choice, from the selection of quantized models (Qwen3.5-9B-IQ4\_XS.gguf) to the
decision to use deterministic scoring alongside probabilistic LLM output.
\section{Conclusion}
\label{sec:conclusion}
We presented NSIDDx, a neuro-symbolic pipeline with explicit negation,
ternary-hybrid scoring, and full-stack override --- and a design
framework arguing that diagnostic AI must prioritise tractability of
disagreement over raw accuracy. Evaluation characterises three failure
modes of the LLM+RAG paradigm under open-ended conditions (vocabulary
mismatch, confidence-without-grounding, negation inversion) that make
clinician oversight a routine requirement, not an edge case.
Offline deployability on consumer hardware is a prerequisite for
equitable access in the settings this system is designed to serve.
A case study demonstrates the override mechanism. We distill five
design principles for clinician-in-the-loop clinical NLP and invite
the prospective studies needed to validate the claim at scale.
\section*{Limitations}
\label{sec:limitations}

We document the known limitations of NSIDDx alongside the design choices
that partially address each.

\paragraph{Natural language symptom coverage.}
MedSpaCy TargetRule matching performs literal string lookup against a
large synonym dictionary. A semantic fallback (Abhinand/MedEmbed-large-v0.1, cosine threshold~0.7)
resolves colloquial terms such as \textit{``SOB''} to dyspnea, but the
threshold was not ablated and false positives are possible; match quality
is uncertain.
Each extracted symptom is tagged with its resolution method (literal/semantic);
the practitioner can inspect and correct any mapping via
\texttt{/add\_symptom}.

\paragraph{PrimeKG phenotype coverage.}
PrimeKG has limited coverage of common diseases: many conditions a general
practitioner would recognise confidently (e.g., pharyngitis, bronchitis)
lack curated phenotype entries in the graph, producing zero or low KG
scores for diagnoses that are clinically straightforward. The HPO/MONDO
matrix score is unaffected by this gap. NSIDDx is most valuable for
complex, multisystem presentations where GPs genuinely benefit from
structured support. This limitation is partially mitigated by the
dual-source design: while PrimeKG scores may be near-zero for common
conditions, the HPO/MONDO matrix score remains operational, and the
LLM-generated DDx is not suppressed by low KG scores.

\paragraph{LLM hallucination in DDx reasoning.}
The LLM may generate plausible but factually incorrect reasoning chains,
particularly for rare diseases with sparse training data representation. The
symbolic scoring layer exists precisely as an independent validation step:
if the LLM proposes a diagnosis that the matrix score actively contradicts,
the practitioner sees both the narrative reasoning and the symbolic
contradiction. The system does not suppress the LLM's output; it presents
it alongside its symbolic assessment.

\paragraph{Absent symptom reliability.}
The pipeline treats clinician-reported negations as structurally meaningful
diagnostic signals. In real settings, patients may not
volunteer absent symptoms unless explicitly asked. The history-taking agent partially addresses this through targeted pertinent-negative elicitation, but cannot guarantee completeness.

\paragraph{Single-language support.}
The pipeline currently operates on English clinical text. Many low-resource
settings operate in Hindi, Swahili, or other languages. Cross-lingual
extension is outside the scope of this version; the modular architecture is
designed to support alternative extraction front-ends.

\paragraph{Single-encounter reasoning.}
NSIDDx reasons over a single clinical encounter. Chronic or evolving
presentations with important longitudinal signal are outside the current
design scope.

\paragraph{Phenotype modifier terms in scoring.}
The graph correctly captures qualifier relationships between HPO terms via
PrimeKG's \texttt{phenotype\_modifier\_of} edges. However, the symbolic
scoring layer does not exploit this binding: modifier terms extracted by
MedSpaCy are entered into $\mathbf{V}_p$ as independent $+1$ entries,
creating a gap between the graph's semantic fidelity and the scoring
layer's representational granularity. Composing modifier-symptom pairs into
single weighted phenotype entries prior to scoring is a direction for future
work.

\paragraph{Evaluation scope.}
Evaluation is conducted on 750 cases from the CUPCase benchmark across two
cohorts --- real-world published case reports. The sample reflects hardware
constraints (one to three minutes per case on the target hardware). Notably, CUPCase cases are
selected for their diagnostic interest and completeness, which may
overestimate system performance compared to routine clinical notes that
are often fragmented or incomplete. Real-world EHR validation and
evaluation on the full CUPCase corpus remain necessary next steps before
clinical deployment.

\paragraph{Semantic matching permissiveness.}
The evaluation relies on semantic similarity (cosine similarity of sentence
embeddings) as a matching metric. Across the CUPCase cohort, 83.5\% of rare
scanner outputs and 41.6\% of DDx@5 semantic hits carry no tokenically
verifiable relationship to the ground-truth diagnosis. Whether these
semantic hits represent genuine synonymy or embedding artefacts cannot be
determined without expert review --- a core motivation for the practitioner
override interface.

\paragraph{Human-in-the-loop demonstration.}
The override interventions in Section~\ref{sec:case} were performed by a
researcher reviewing the case narrative alongside the automated symptom
vector, not by a clinician in a live diagnostic setting. The case study
demonstrates the mechanism functions as designed but provides no
evidence about human behaviour with the system. Prospective user studies
with practising clinicians are necessary and explicitly invited.

\paragraph{Scoring formula ablation.}
The ternary-hybrid scoring formulas were designed for interpretability
rather than diagnostic accuracy. No hyperparameter tuning was performed;
the equal-weight combination (0.5/0.5) was selected for transparency. Future
work should explore learned weights that preserve interpretability while
improving discrimination.

\section*{Ethical and Societal Implications}
\paragraph{Positive Impact Definition.}
For us, positive impact in clinical NLP is not primarily measured in
benchmark accuracy. It is measured in \textit{access}: whether a
practitioner in a clinic with no specialist referral network, no reliable
internet connection, and no clinical decision support infrastructure can use
a system to surface a differential diagnosis they might otherwise have
missed, inspect the reasoning behind it, and make a better-informed decision
for their patient. NSIDDx is designed with this practitioner in mind. Its
offline deployability, consumer hardware requirements, and
practitioner-in-the-loop architecture are not technical constraints — they
are the design goal. Following the NLP4PI workshop's emphasis on impact
grounding \citep{pant-etal-2025-health}, we define positive impact as
democratized access to structured differential reasoning --- the ability of
any practitioner, regardless of institutional resources, to receive a
transparent, auditable diagnostic aid that respects clinical expertise and
local data sovereignty.
\paragraph{Assistive-Only Design Principle.}
NSIDDx is an assistive tool, not an autonomous diagnostic system. It does
not make diagnoses. Every output — the DDx ranking, the PL string, the
phenotypic graph — is presented to the practitioner as structured
information to reason with, not a decision to accept. The system is
explicitly designed so that disagreement is easy: the practitioner can add
symptoms, remove candidates, and rerun the pipeline in seconds.
\paragraph{Automation Bias and Over-Reliance Risk.}
Any decision support system carries a risk of automation bias: the tendency
of clinicians to anchor on system outputs even when their own clinical
judgment diverges. The PL strings and phenotypic graphs are designed to
\textit{invite} disagreement — a practitioner who reads
$(\text{Alopecia}) \wedge (\text{Malar rash}) \Rightarrow \text{SLE}$ and
knows the patient also has a finding not captured in the system has a clear
signal that the system's evidence base is incomplete. Nevertheless, we
acknowledge that the risk of over-reliance cannot be fully designed away and
should be addressed through practitioner training and deployment guidance.
\paragraph{Rare Disease Scanner and Diagnostic Heuristics.}
We deliberately invert the hoofbeats heuristic for low-resource settings:
when specialist referral is unavailable, surfacing a rare disease candidate
with moderate confidence is more ethical than suppressing it. This inversion
is bounded by three constraints: the scanner is optional and off by default;
it activates only for presentations with eight or more symptoms; and
low-scoring candidates (including negative matrix scores) are visually
deprioritised.
\paragraph{Evaluation on real-world case reports.}
The evaluation uses 750 cases from the CUPCase benchmark across two cohorts,
a publicly available collection of real-world BMC patient case reports.
The sample reflects inference time constraints on consumer hardware. Validation on
institution-specific EHR data, with appropriate ethics approvals and data
governance, is required before any clinical deployment.
\paragraph{Review and Governance.}
NSIDDx is explicitly positioned as a research prototype, not a clinical
tool. Any future deployment would require institutional review board
approval, HIPAA/GDPR compliance audits, and a phased clinical validation
protocol. The system's architecture --- local processing, no cloud
dependency, auditable reasoning chains --- is designed to facilitate, not
circumvent, these governance requirements.
\paragraph{Equity and language access.}
The instantiated system currently operates only in English. This limits immediate
applicability in many of the low-resource settings it is designed to serve.
We recognise this as a significant equity gap and identify multilingual
extension as a priority for future work.
\paragraph{Data Retention and Sovereignty.}
NSIDDx operates entirely locally. No patient information is sent to external
servers or APIs. This is both a practical requirement for offline deployment
and an ethical requirement for patient data sovereignty in settings where
data protection infrastructure may be limited.

\bibliography{custom}

\appendix
\section{Sarcoidosis Case: Negation-Retaining Phenotype Graph}
\label{sec:sarc_transcripts}
The following is the Negation-Retaining Phenotype Graph output for the
corrected Sarcoidosis demonstration case from Section~\ref{sec:case},
showing present and absent symptom nodes with their HPO identifiers and
PrimeKG-derived \texttt{explains} edge weights for each DDx candidate.
\begin{footnotesize}
\noindent\textbf{Patient -- present (7):}
\begin{quote}
dyspnea (HP:0002094)\\
hilar lymphadenopathy (HP:0034388)\\
mediastinal lymphadenopathy (HP:0100721)\\
pleural effusion (HP:0002202)\\
subcutaneous nodule (HP:0001482)\\
granuloma (HP:0032252)\\
splenomegaly (HP:0001744)
\end{quote}
\noindent\textbf{Patient -- denies (1):}
\begin{quote}
palpitations (HP:0001962)
\end{quote}
\noindent\textbf{Candidates (green nodes):}
\begin{quote}
Sarcoidosis (ORPHA:797): explains dyspnea (w=3.739),
pleural effusion (w=3.317), subcutaneous nodule (w=3.165),
mediastinal lymphadenopathy (w=2.781); also linked to
$\neg$palpitations via ``denies''\\
Granulomatosis with polyangiitis: explains pleural effusion
(w=3.317), subcutaneous nodule (w=3.165); narrower coverage\\
Lung TB: no resolved HPO overlap, disconnected node with
zero scores\\
NHL: no resolved HPO overlap, disconnected node\\
Pulmonary fungal disease: no resolved HPO overlap,
disconnected node
\end{quote}
\end{footnotesize}

\section{Evaluation Data}
\label{sec:evaldata}

The evaluation set consists of 750 cases drawn from the CUPCase benchmark
\citep{cupcase2025} across two cohorts: a 500-case random sample
representing the full distribution of clinically uncommon presentations, and
a 250-case exact-match sample representing the upper bound of ontology
coverage within CUPCase. The full 3,562-case corpus remains unevaluated due
to inference time constraints on consumer hardware.

The complete filtered dataset, together with all evaluation scripts and
code used in this paper, is publicly available at
\url{https://github.com/joetheguide2/NSIDDX-}. The raw per-case result
files backing this evaluation are \texttt{part\_1\_qwen9b\_semantic.csv}
(CUP, 500-case cohort) and \texttt{exact\_qwen9b\_semantic.csv} (WELL,
250-case cohort). The repository also contains earlier exploratory runs
and threshold variants retained for transparency; these do not reflect
the final reported numbers, which are the two files named above. Some
summary documents in the repository (e.g. disease/symptom counts in the
top-level README) report raw, pre-filter counts rather than the corrected,
filtered counts used in the scoring pipeline and reported in Section~3.

\section{Scoring Mechanism Detail}
\label{sec:scoring}

The scoring engine produces four scores per candidate disease, displayed as
\texttt{HPO\_m} / \texttt{KG\_m} (matrix) and \texttt{HPO\_h} /
\texttt{KG\_h} (hybrid). Both formulas are applied to both sources.
Table~\ref{tab:scoring} provides an interpretation guide for reading score
combinations in clinical practice.

\begin{table}[ht]
\small
\centering
\setlength{\tabcolsep}{4pt}
\begin{tabular}{@{}p{1.2cm}p{1.2cm}p{4.6cm}@{}}
\toprule
\textbf{Matrix} & \textbf{Hybrid} & \textbf{Clinical interpretation} \\
\midrule
High & High & Strong match; disease fits both positive findings and full expected profile \\
\addlinespace
Low & High & Disease explains findings but patient is missing symptoms the disease requires \\
\addlinespace
Negative & Any & Active contradiction; inspect grey nodes in graph for specific conflicts \\
\addlinespace
Any & Low & Weak explanatory fit; disease does not account for the patient's symptom cluster \\
\addlinespace
Zero & Zero & No phenotype overlap; consider \texttt{/rare\_disease\_scan} \\
\addlinespace
HPO high, KG low & --- & KG resolution likely inaccurate; check \texttt{KG} $\rightarrow$ mapping \\
\bottomrule
\end{tabular}
\caption{Four-score interpretation guide. HPO/KG discrepancies surface
  knowledge graph noise and are flagged for the practitioner.}
\label{tab:scoring}
\end{table}

\end{document}